\documentclass[12pt,a4paper]{article}

\usepackage{textcomp}
\usepackage{amsmath}
\usepackage[linesnumbered,ruled]{algorithm2e}
\usepackage[titletoc]{appendix}
\usepackage[authoryear,numbers,sort&compress]{natbib}
\usepackage{multicol}
\usepackage{color}
\usepackage{tabularx}
\usepackage{authblk}
\usepackage{subfig}
\usepackage{amscd, amssymb, tikz}
\usepackage[margin=2cm,includehead, includefoot]{geometry}
\usepackage{hyperref}
\usepackage{float}
\usepackage{wasysym}
\usepackage{pifont}% http://ctan.org/pkg/pifont
\usepackage{fancyvrb}
\usepackage{mathrsfs}
\usepackage{verbatim,color,lipsum}
\usepackage{marvosym}
\usepackage{setspace}
\usepackage{graphicx,etoolbox}

\makeatletter
\patchcmd{\Gin@setfile}% <cmd>
{\ProvidesFile}% <search>
{\xdef\imgfilename{#3}\ProvidesFile}% <replace>
{}{}% <success><failure>
\makeatother

\newcommand{\bs}[1]{\mathbf#1}
\newcommand{\bsy}[1]{\boldsymbol#1}
\newcommand{\intd}{\textnormal{d}}

\newcommand{\KL}[2]{\text{KL}\left(#1 \mid\mid #2\right)}
\newcommand{\E}[2]{\text{E}_{#1}\left[#2\right]}

\newcommand{\enc}[1]{\textbf{e}\left(#1\right)}

\newcommand{\encgam}[1]{\text{e}_\gamma(#1)}
\newcommand{\dec}[1]{\textbf{d}\left(#1\right)}

\newcommand{\sige}[2]{\sigma^2_{\text{e,}#1}(#2)}
\newcommand{\Sige}[1]{\bsy{\Sigma}_{\text{e}}(#1)}

\newcommand{\Sigd}[1]{\bsy{\Sigma}_{\text{d}}(#1)}
\newcommand{\bi}{\bs{b}_i}

\renewcommand{\xi}{\bs{x}_{i}}

\newcommand{\zi}{\bs{u}_i}

\newcommand{\appropto}{\mathrel{\vcenter{
			\offinterlineskip\halign{\hfil$##$\cr
				\propto\cr\noalign{\kern2pt}\sim\cr\noalign{\kern-2pt}}}}}

\newcommand{\hang}%
    {\hangindent\parindent}
\newcommand{\textindent}[1]%
    {\indent\llap{\hbox to \parindent{#1\hfill}}\ignorespaces}
\newcommand{\litem}%
    {\par\hang\textindent}
\newcommand{\litemitem}%
    {\par\indent\hangindent2\parindent\textindent}
\newcommand{\noparlitem}%
    {\hangindent\parindent\textindent}
\newcommand{\noparlitemitem}%
    {\hangindent2\parindent\textindent}
\newcommand{\litemitemitem}%
    {\par\indent\indent\hangindent3\parindent\textindent}
  
\def\correspondence#1{\gdef\@corresp{#1}}

\begin{document}

\pagenumbering{arabic}

%-------------------------------------------------------------------------- 

\title{\bf Emulation of greenhouse-gas sensitivities using variational autoencoders}

% \author[affil]{given_name}{surname}

\author[1]{\small Laura Cartwright}
\author[1]{Andrew Zammit-Mangion}
\author[2]{Nicholas M. Deutscher}

\affil[1]{School of Mathematics and Applied Statistics, University of Wollongong, Wollongong, Australia}
\affil[2]{Centre for Atmospheric Chemistry, School of Earth, Atmospheric and Life Sciences, University of Wollongong, Wollongong, Australia}

%% The [] brackets identify the author with the corresponding affiliation. 1, 2, 3, etc. should be inserted.

\date{}

\maketitle

\begin{center}
\small \textbf{Correspondence:} Laura Cartwright (lcartwri@uow.edu.au)
\end{center}

\vskip 15pt

\begin{abstract}

	Flux inversion is the process by which sources and sinks of a gas are identified from observations of gas mole fraction. The inversion often involves running a Lagrangian particle dispersion model (LPDM) to generate sensitivities between observations and fluxes over a spatial domain of interest. The LPDM must be run backward in time for every gas measurement, and this can be computationally prohibitive. To address this problem, here we develop a novel spatio-temporal emulator for LPDM sensitivities that is built using a convolutional variational autoencoder (CVAE). With the encoder segment of the CVAE, we obtain approximate (variational) posterior distributions over latent variables in a low-dimensional space. We then use a spatio-temporal Gaussian process emulator on the low-dimensional space to emulate new variables at prediction locations and time points. Emulated variables are then passed through the decoder segment of the CVAE to yield emulated sensitivities. We show that our CVAE-based emulator outperforms the more traditional emulator built using empirical orthogonal functions and that it can be used with different LPDMs. We conclude that our emulation-based approach can be used to reliably reduce the computing time needed to generate LPDM outputs for use in high-resolution flux inversions. 
	
\end{abstract}

\section{Introduction}

It is becoming increasingly urgent to not only identify, but also quantify, the contributors to climate change (see, e.g., \citealp{annan_2002}; \citealp{flannery_2009}; \citealp{zidansek_2009}; \citealp{reddy_2010}). The major contributor to the enhanced greenhouse effect is the emission of gases that have the capacity for trapping heat in the atmosphere. Methane (CH$_4$) is the second most prevalent anthropogenic greenhouse gas (GHG) in the atmosphere after carbon dioxide (CO$_2$); however, by mass it is much more potent than CO$_2$ (1 metric tonne of CH$_4$ is approximately 28-34 times more effective at trapping energy than 1 metric tonne of CO$_2$ over a hundred year period \citep{Myhre_2013}). Therefore, identifying and quantifying GHG emissions, especially CH$_4$ emissions, is of considerable interest to atmospheric scientists and policy makers alike. This is, however, a difficult problem: while measurements of CH$_4$ mole fraction can be done reasonably accurately using sophisticated instruments, estimating the spatial distribution of sources of CH$_4$ from those measurements is an ill-posed problem. 

\vskip 10pt

Flux inversion is at the core of many methods for quantifying GHG emissions (recent examples include %\citealp{gurney_2002}; 
\citealp{humphries_2012}; \citealp{luhar_2014}; \citealp{Alexe_2015}; \citealp{ganesan_2015}; \citealp{zammit_2015}; \citealp{houweling}; \citealp{lucas_2017}; \citealp{Bergamaschi_2018}; \citealp{Cressie_2018}; \citealp{feitz_2018}; \citealp{cartwright_2019}). It can be summarised as the process of comparing observed GHG mole fractions to those simulated using an atmospheric transport model (ATM), in order to estimate the fluxes that are  contributing to the observed mole fractions. Often, this process involves ``backward simulation", where one simulates particle trajectories at an observation location backward in time through an ATM. The derived relationship between each observation and the flux from this simulation is a \emph{sensitivity} that evolves in both space and time. These sensitivities can be time-integrated over the duration of the simulation, to yield a \emph{sensitivity plume} that identifies the sensitivity of a specific observation to the (here, assumed time-invariant) flux field. These sensitivity plumes can then be used in a Bayesian inference framework to locate and quantify the sources or sinks \cite[e.g.,][]{zammit_2016}. 

\vskip 10pt

Many highly sophisticated ATMs exist, and in this work we consider a family of models called Lagrangian Particle Dispersion Models (LPDMs). Some examples of LPDMs are the Numerical Atmospheric Modelling Environment \citep[NAME,][]{name}, the FLEXible PARTicle Dispersion Model \citep[FLEXPART,][]{flexpart_104}, and the Stochastic Time-Inverted Lagrangian Transport Model \citep[STILT,][]{stilt}. LPDMs are used to simulate atmospheric dispersion of particles moving through the air. Trajectories are computed based on the Brownian motion of the particles \citep[see][Chapter 9, for a thorough explanation]{devisscher_2013}. LPDMs involve a large number of parameters (often well over 50), and though highly representative of the physical environment, their large computational cost can be prohibitive in flux inversion applications \citep{francom_2019}. This can lead, for example, to flux inversions being performed over coarser spatio-temporal resolutions than desired, restricting one from resolving fine-scale components of interest. 

\vskip 10pt

The computational burden associated with running LPDMs has led to the investigation of ways in which they can be used to inform the construction of surrogate models. For example, \cite{lucas_2017} use decision/regression trees to predict outputs from a hybrid of the FLEXPART and Weather Research and Forecasting (WRF) models, FLEXPART-WRF. Over small (local-scale) regions, simpler models such as the Gaussian plume dispersion model are often used (e.g., \citealp{Borysiewicz_2012}; \citealp{feitz_2018}; \citealp{cartwright_2019}). While much less sophisticated than an LPDM, the Gaussian plume dispersion model is quick to evaluate, allowing it to be easily calibrated (a procedure in which numerical model parameters are adjusted so that the simulated concentrations are a good fit to measurements) during the inversion. However, such an approach cannot be reliably used over regions larger than a few square kilometres, since the assumptions of the Gaussian plume dispersion model (e.g., constant windspeed and constant vertical Eddy diffusion) are not reasonable over large distances \citep{riddick_2017}. 

\vskip 10pt

\begin{figure}[t!]
	\centering \includegraphics[scale = 0.55]{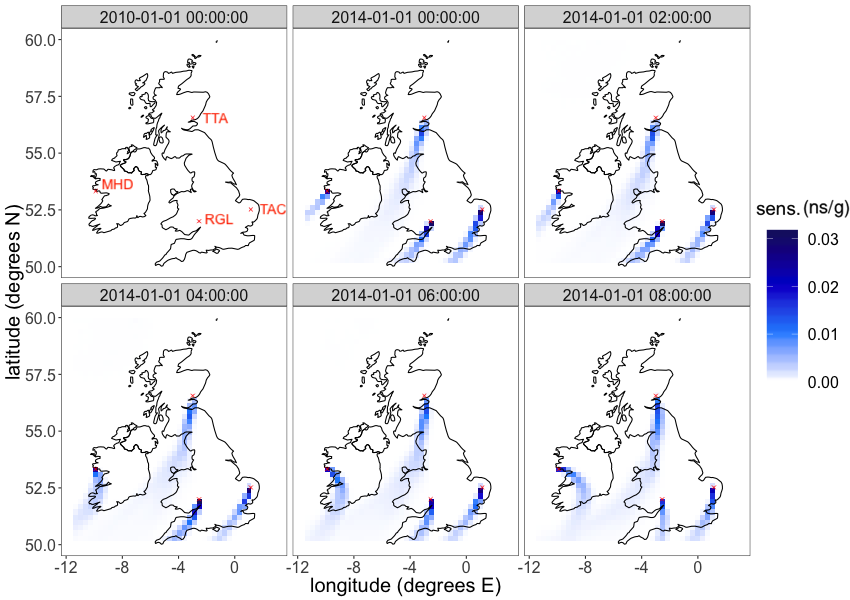}
	\caption{Top-left panel: Four methane measurement sites from the UK and Ireland, located at Mace Head (MHD), Ridge Hill (RGL), Tacolneston (TAC), and Angus (TTA). Other panels: The sensitivity of mole fraction to the flux over two-hourly time intervals on 1 January 2014, based on a release from each of the four sites, as output from the Numerical Atmospheric Dispersion Modeling Environment (NAME). Here ``sens." represents the maximum sensitivity at each grid cell, where the maximum is taken over each of the four NAME outputs from the corresponding time point (one per measurement site).}
	\label{fig:plumes-corr}
\end{figure}

The core idea of emulating LPDM output is not new in this article. For example, \cite{Harvey_2018} present a polynomial-based model, while \cite{francom_2019} use Bayesian multivariate adaptive regression splines as the emulator basis. In the former case, the emulator is used directly on a subset of inputs to the LPDM, while in the latter, the emulator also takes information from the principal components of simulated outputs, using empirical orthogonal functions (EOFs). However, in both cases, the emulated values represent atmospheric pollutant mole fractions downwind from a \emph{single} point source, at a small number of locations and time points during the simulation. Further, measurements of the atmospheric pollutant (volcanic ash, for example) used in these works are unlikely to be affected by other sources. Our more general case is different, as measurements of methane are likely to be affected by multiple sources, and we need to consider a flux field consisting of many (not necessarily known) sources and sinks. To accommodate these requirements, we propose ``borrowing strength" from sensitivity plumes obtained via an LPDM at a small number of locations in order to construct additional plumes at new locations. That is, we extract information about the shape/structure of plumes, and then use this information to emulate the LPDM output at additional spatio-temporal locations. The motivation for this can be seen in Figure \ref{fig:plumes-corr}, where we note that sensitivity plumes for observations of methane that are nearby in space and time appear to be highly correlated, likely due to the smoothly varying nature of the governing meteorological influences (e.g., wind speed, wind direction, air pressure). Our approach allows us to construct time-integrated methane sensitivity plumes over the entire spatio-temporal domain of interest with relative ease.

\vskip 10pt

While it is possible to build an emulator that predicts the LPDM sensitivity at each individual cell in the model grid, it would be difficult to construct in practice, since popular emulators such as the Gaussian process (GP) emulator \citep[see][for example]{rasmussen_2006} do not scale well with the dimensionality of the parameter space. We instead propose a two-fold approach. First, we extract ``important features" of the LPDM-built sensitivity plumes, and use these features to reduce the dimension of the output down to a small number, with each dimension representing a prominent feature of the sensitivity plumes. Second, we emulate variables associated with these features at new spatio-temporal locations. By applying a reverse transformation on the features we then produce predicted LPDM-sensitivity-plumes on the original space. 

\vskip 10pt 

For the first, dimension-reduction, step, a number of methods are available, including principal component analysis (PCA) or EOF analysis (e.g., \citealp{higdon_2008}; \citealp{zhang_2018}; \citealp{francom_2019}), gradient-based kernels \citep[e.g.,][]{liu_2017}, sliced inverse regression \citep[e.g.,][]{li_1991}, and variable selection algorithms (e.g., \citealp{wolf_2005}; \citealp{Harvey_2018}). Machine learning methods are also popular for dimension reduction. When carrying out dimensionality reduction where reconstruction quality of inputs is important, in machine learning one often considers the autoencoder \citep[see][chapter 14]{goodfellow_2016}. An autoencoder first reduces the dimension of inputs by mapping them, via an encoder, onto a latent space. Via a decoder, it then takes values on the latent space and constructs an approximation to the original input via a reverse mapping. In this article we propose the use of a convolutional variational autoencoder (CVAE); see \citet{kingma_2014} for a full introduction to variational autoencoders. The ``variational" portion of the name arises from the variational-Bayes-based probabilistic model used to build the autoencoder, which allows us to avoid over-fitting and increase the generalisability of the fitted model. 

\vskip 10pt

For the second, emulation, step, we use spatio-temporal GP emulators on the latent space, which allow us to quantify the uncertainty arising from not running the LPDM at new spatio-temporal locations of interest. We see improvement in the emulated plumes, both visually and in terms of mean-squared error, when the CVAE is used for dimensionality reduction instead of EOFs.

\vskip 10pt

The remainder of the article is organised as follows. In Section 2, we describe the methodology behind EOF analysis and the CVAE, and give a brief description of the GP emulators we employ. In Section 3, we present and discuss the results we obtain when emulating the sensitivity plumes from two LPDMs. Section 4 concludes.

\section{Methodology} \label{sec:methodology}

Consider a rectangular spatial domain $D \subset [-180, 180]^{\circ} \times [-90, 90]^{\circ}$. Let $\bs{s} \equiv (s_1, s_2)'$ be a spatial location in $D$ with units of degrees, and $\bs{w} \equiv (\bs{s}'; t)'$, where $t$ denotes time in UTC. Define $Y_m(\bs{w})$ to be the spatio-temporal methane mole fraction process at $\bs{w}$, and $Y_f(\bs{s})$ to be the (assumed time-invariant) spatial methane flux process at $\bs{s}$. The \emph{sensitivity} of $Y_m$ to $Y_f$  depends on a bivariate function $b(\cdot, \cdot)$, and is given by \citep[][]{zammit_2015},

\begin{equation}
Y_m(\bs{w}) = \int_{D} b(\bs{w}, \bs{r}) Y_f(\bs{r})\intd \bs{r}.
\label{eq:sens}
\end{equation}

In practice, the integral in \eqref{eq:sens} can rarely be done in closed form, and is instead approximated using an ATM (such as an LPDM in this case). Specifically, the sensitivity is evaluated numerically over a fine space-time grid.

\vskip 10pt

Let $D^G$ be a rectangular spatial gridding of $D$, consisting of $K$ equally-sized grid cells, and let $\bs{s}_k \equiv (s_{k,1}, s_{k,2})',$ $k \in 1, 2, \dots, K,$ be the lower-left corner of the $k$th grid cell in $D^G$, where $s_{k,1}$ corresponds to degrees longitude, and $s_{k,2}$ to degrees latitude. %Let $Y_f(\bs{s}_k)$ be the flux process at $\bs{s}_k$. 
We are interested in how the spatio-temporal mole fraction process $Y_m(\cdot)$ at a collection of $N$ points, $\bs{w}_1, \bs{w}_2, \dots, \bs{w}_N$, relates to the spatial flux process, $Y_f(\cdot)$, where $\bs{w}_i \equiv (\tilde{\bs{s}}'_i; t_i)'$, $\tilde{\bs{s}}_i \equiv (\tilde{s}_{i,1}, \tilde{s}_{i,2})' \in D$, and $t_i$ is a time point. Following the discretisation of (\ref{eq:sens}), we have the linear relationship 

\begin{equation}
\bs{Y}_m = \bs{B}\bs{Y}_f,
\end{equation}  

where $\bs{Y}_m \equiv \left(Y_m(\bs{w}_i) \colon i = 1, 2, \dots, N\right)',$ and $\bs{Y}_f \equiv \left(Y_f(\bs{s}_k) \colon k = 1, 2, \dots, K\right)'$. The $i$th row of $\bs{B}$, denoted $\bs{b}_i$, describes the sensitivity of $Y_m(\bs{w}_i)$ to the flux at all $\bs{s}_k, k = 1, 2, \dots, K$. We determine each $\bs{b}_i$ using an LPDM.

\vskip 10pt

Suppose now that we have an additional set of $M$ points, $\bs{w}^*_1, \bs{w}^*_2, \dots, \bs{w}^*_M$, at which we wish to model the mole fraction in response to $\bs{Y}_f$, where for $j = 1, 2, \dots, M$, $\bs{w}^*_j \equiv (\tilde{\bs{s}}_j^{*'}; t^*_j)$, $\tilde{\bs{s}}^*_j\equiv (\tilde{s}^*_{j,1}, \tilde{s}^*_{j,2})'$ is a point in $D$, $t^*_j$ is a point in time, $t_1 \leq t^*_1,$ and $t_N \geq t^*_M.$ The primary aim of this work is to construct an emulator that can be used to construct a new matrix $\widehat{\bs{B}}^*$ from $\bs{B}$,  where each row, $\widehat{\bs{b}}^*_j$, is the \textit{predicted} sensitivity of $Y_m(\bs{w}^*_j)$ to the flux at all $\bs{s}_k \in D^G$. The construction of such an emulator is the subject of the following sections.

\subsection{Plume feature extraction} \label{sec:feat}

Before constructing the emulator, we first focus on reducing the dimension of the plumes. We do this since emulating a $K$-dimensional quantity for large $K$ is difficult, and also wasteful due to sensitivity vectors containing a lot of structure. As can be seen from Figure \ref{fig:plumes-corr}, much of the sensitivity over $D^G$ is often zero, and the ``important information" about the sensitivity of $Y_m(\bs{w}_i)$ is contained in only a small subset of $\bs{b}_i$. This suggests that we could capture the important information regarding the shape (structure) and intensity of these plumes through a small set of features.

\vskip 10pt

We now describe two methods for feature extraction/dimension reduction, one using EOFs, and one using a CVAE. 

\subsubsection{Empirical orthogonal functions}

EOF analysis \citep{halldor_1997} refers to the process of decomposing the spatio-temporal data into a set of orthogonal basis vectors that can help describe the spatial structure of the data. EOF analysis is used to extract spatial features from data indexed over time, and hence can be thought of as a spatio-temporal version of principal component analysis \citep[PCA,][Chapter 2]{wikle_2019}. The basis vectors are constructed successively in such a way that each is orthogonal to those constructed previously, and in a way that they account for as much spatial variability in the data as possible. That is, the first basis vector explains the largest amount of spatial variability in the data, while the second basis vector explains the second largest amount of spatial variability, and so on. 

\vskip 10pt

We derive the EOFs via the singular value decomposition (SVD) of $\bs{B}$, which is of size $N \times K$. The SVD of a matrix has three components, one of which contains the singular values of the matrix. Specifically, when $N > K$,

\[\bs{B} = \bs{U}\bs{D}\bs{V}',\]

where $\bs{U}$ is an $N \times K$ orthonormal matrix of left singular vectors, $\bs{D}$ is a $K \times K$ matrix with the singular values along its main diagonal, and $\bs{V}$ is a $K \times K$ orthonormal matrix of right singular vectors. It can be easily shown that the columns of $\bs{V}$ are the $K$ eigenvectors of $\bs{B}' \bs{B}$, and that the diagonal of $\bs{D}$ contains the square-root of the eigenvalues. The product $\bs{D}\bs{V}'$ makes up the EOFs, and each column in $\bs{U}$ represents the coefficients for one EOF. 

\vskip 10pt 

We construct an approximate representation of $\bs{B}$ using the first $r$ EOFs, by computing 

\begin{equation}
\bs{B} \approx \bs{U}_{r} \bs{D}_{r} \bs{V}_{r}', \label{eq:udv-construct}
\end{equation}

where $\bs{U}_{r},$ and $\bs{V}_{r}$ are the first $r$ columns of the matrices $\bs{U}$ and $\bs{V},$ respectively, and $\bs{D}_{r},$ consists of the first $r$ rows and columns of $\bs{D}$. The $i$th row in $\bs{U}_{r}$, denoted $\bs{u}_i$, represents the coefficients for the first $r$ EOFs for the $i$th plume. When using EOFs, we therefore emulate (Section \ref{sec:emulation}) based on the information in the rows of $\bs{U}_r$, to predict sensitivities at new locations. Specifically, for spatio-temporal locations $\bs{w}_1^*, \bs{w}_2^*, \dots, \bs{w}_M^*$, we obtain an $M \times r$ matrix of coefficients through emulation, $\widehat{\bs{U}}_r^*$. We then construct $\widehat{\bs{B}}^*$ (of size $M \times K$) by computing 

$$\widehat{\bs{B}}^{*} = \widehat{\bs{U}}_r^* \bs{D}_r \bs{V}'_r.$$

This approach to emulation is often seen in the computer experiment literature and has been shown to be successful in various application settings \citep[e.g., ][]{zhang_2018, higdon_2008, holden_2015}.

\subsubsection{Convolutional variational autoencoder}

While EOFs are widely used in the atmospheric sciences, they can also be restrictive. For example, \cite{hannachi_2007} notes that the spatial orthogonality of EOFs restricts their ability to describe localised patterns within the domain, and can lead to difficulty in pattern interpretation as well as domain-dependence issues. In addition, the SVD yields a linear decomposition; as a result, underlying non-linear relationships between the ``inputs" and their ``features" could be overlooked \citep{bishop_2000}. In this section we explore the use of a deep learning architecture to extract the basis functions with which to do emulation. A deep learning model is a collection of simple, interconnected layers of ``neurons", often used to extract important features from a dataset. Neural networks feature a collection of weights that are optimised according to one or more criteria during the training of the network \citep[][Chapter 1]{Gurney_1997}. In a fully-connected neural network (i.e., when each neuron in one layer is connected to all neurons in the next layer) for image data, a very large number of weights are required. As images have a known, gridded structure, it is possible to use convolutional layers instead, which are not fully connected \citep{Soydaner_2019}. In a convolutional neural network (CNN), some or all of the layers use a convolution in place of the standard matrix multiplication used in fully-connected neural networks. 

\vskip 10pt

In practice, we often not only wish to find features of an image, but also to use these features to reconstruct the original image. The architecture used for this purpose is known as a \textit{convolutional autoencoder}, which can be thought of as a two-piece CNN: an encoder, and a decoder. The encoder maps each image onto a latent space of variables $\bs{u}$, which represent the value of the extracted features. Then, these features can be mapped back onto the original space (reconstructing the original image) via the decoder. The CNN autoencoder has been used for prediction elsewhere \citep[e.g.,][]{xu_2020}, however it does not require that the latent space be contiguous \citep{song_2019}. This can be problematic if one wishes to use the decoder for emulation, as in our application, since, as with most emulators, we rely on the output varying smoothly with the input when predicting at a new location.

\vskip 10pt

The \textit{variational autoencoder} \citep[VAE,][]{kingma_2014} differs from the standard autoencoder in that the latent variables are no longer deterministic, but are now random variables. In addition to the mean squared error, the CVAE objective function also includes a Kullback-Leibler divergence (KL-divergence), which penalises for (and, in practice, does not allow)  discontinuities in the latent space. 

\vskip 10pt 

Consider $N$ vectorised training images $\bs{b}_1, \bs{b}_2, \dots, \bs{b}_N,$ each of dimension $K \times 1$, and an $r$-dimensional latent variable $\zi \equiv (u_{i, 1}, u_{i, 2}, \dots, u_{i, r})'$ for each $\bi, i = 1, 2, \dots, N$. In a CVAE, we seek to maximise a lower bound for the marginal likelihood $p(\bi ; \bsy{\theta}),$ where $\bsy{\theta} \equiv (\bsy{\theta}_e', \bsy{\theta}_d')',$ and $\bsy{\theta}_e$ and $\bsy{\theta}_d$ are the encoder and decoder parameters, respectively, that need to be estimated. In particular, we have that 
\begin{align} 
p(\bi ; \bsy{\theta}) &= \int p(\bi, \zi ; \bsy{\theta})\ \intd \zi \nonumber \\
&= \int q(\zi \mid \bi; \bsy{\theta}_e) \frac{p(\bi \mid \zi; \bsy{\theta}_d) p(\zi)}{q(\zi \mid \bi; \bsy{\theta}_e)}\ \intd \zi, \nonumber \\
\intertext{where $q(\zi \mid \bi; \bsy{\theta}_e)$ is the approximation to the posterior distribution $p(\zi \mid \bi, \bsy{\theta}_d)$ when making inference using variational Bayes. By Jensen's inequality, we therefore have that} 
\log(p(\bi; \bsy{\theta}_e))  &\geq \int q(\zi \mid \bi; \bsy{\theta}_e) \log{\left(\frac{p(\bi \mid \zi; \bsy{\theta}_d)p(\zi)}{q(\zi \mid \bi; \bsy{\theta}_e)}\right)} \ \intd \zi \nonumber \\
&= \E{q(\zi \mid \bi; \bsy{\theta}_e)}{\log(p(\bi \mid \zi; \bsy{\theta}_d))} - \KL{q(\zi \mid \bi; \bsy{\theta}_e)}{p(\zi)} \label{eq:negELBO}.
\end{align}

Maximising \eqref{eq:negELBO} is equivalent to minimising
\begin{equation}  
- \E{q(\zi \mid \bi; \bsy{\theta}_e)}{\log(p(\bi \mid \zi; \bsy{\theta}_d))} + \KL{q(\zi \mid \bi; \bsy{\theta}_e)}{p(\zi)}. \label{eq:ELBO}
\end{equation}

We let $\zi \sim \textnormal{Gau}(\bs{0}, \mathbb{I}_{r})$ and $q(\zi \mid \bi; \bsy{\theta}_e) = \textnormal{Gau}\left(\enc{\bi; \bsy{\theta}_e}, \Sige{\bi; \bsy{\theta}_e}\right), i = 1, 2, \dots, N$, where $\enc{\cdot; \bsy{\theta}_e} \equiv \left( e_1(\cdot; \bsy{\theta}_e), e_2(\cdot; \bsy{\theta}_e), \dots, e_r(\cdot; \bsy{\theta}_e)\right)'$ is the encoder mean, and

\[\Sige{\bi; \bsy{\theta}_e} \equiv \text{diag}\left(\sige{1}{\bi; \bsy{\theta}_e}, \sige{2}{\bi; \bsy{\theta}_e}, \dots, \sige{r}{\bi; \bsy{\theta}_e}\right),\] 

is the (diagonal) encoder covariance matrix, with $\sige{\gamma}{\bi; \bsy{\theta}_e}, \gamma = 1, 2, \dots, r,$ nonnegative functions of $\bi$. Under these assumptions, the second term on the right-hand side of \eqref{eq:ELBO} becomes

\vskip -15pt

\begin{align*} 
\KL{q(\zi \mid \bi; \bsy{\theta}_e)}{p(\zi)} &= -\frac{1}{2} \sum_{\gamma = 1}^{r} \left(  \log(\sige{\gamma}{\bi; \bsy{\theta}_e}) + 1 - \sige{\gamma}{\bi; \bsy{\theta}_e} - \encgam{\bi; \bsy{\theta}_e}^2 \right).
\end{align*}

Therefore, under the Gaussian prior and the chosen approximate form of the posterior distribution, the KL-divergence between the approximate posterior distribution and the prior on $\zi$ can be calculated using the encoder-determined mean and variance (in practice, log-variance) for each $u_{i,\gamma}$ \citep{kingma_2014}.

\vskip 10pt

For the first term on the right-hand side of \eqref{eq:ELBO}, we assume a Gaussian likelihood, 

\[p(\bi \mid \zi; \bsy{\theta}_d) = \textnormal{Gau}(\dec{\zi; \bsy{\theta}_d}, \Sigd{\zi}),\]

where $\dec{\cdot; \bsy{\theta}_d}$ is the decoder mean, and $\Sigd{\zi}$ is the decoder covariance matrix. Thus, we assume that the reconstructed images are corrupted with Gaussian noise which, for simplicity, we assume is homoscedastic and uncorrelated; that is, we let $\Sigd{\zi} = c\mathbb{I}_{M}, c > 0$. Now, minimisation of 

\[- \E{q(\zi \mid \bi; \bsy{\theta}_e)}{\log(p(\bi \mid \zi; \bsy{\theta}_d))}\]

cannot be done analytically since $\dec{\cdot; \bsy{\theta}_d}$ is an arbitrary non-linear function of its arguments. We therefore approximate this expectation using Monte Carlo. In particular, 

\begin{equation} 
	- \E{q(\zi \mid \bi; \bsy{\theta}_e)}{\log(p(\bi \mid \zi; \bsy{\theta}_d))} \approx
	\frac{1}{L} \sum_{l = 1}^L \frac{1}{2c} 
	\left\Vert \bi - \dec{\bs{u}_i^{(l)}; \bsy{\theta}_d} \right\Vert_2^2 + \text{const.},
	\label{eq:MC}
\end{equation}

where $\bs{u}_i^{(l)}$ is the $l$th ($l = 1, 2, \dots, L$) random sample from $q(\bs{u}_i \mid \bs{b}_i; \bsy{\theta}_e)$, and ``const." does not depend on the decoder parameters. In our implementation, we therefore train the CVAE by minimising  

\begin{equation}
\sum_{i = 1}^N \left[ \frac{1}{L} \sum_{l = 1}^L  
\left\Vert \bi - \dec{\bs{u}_i^{(l)}; \bsy{\theta}_d} \right\Vert_2^2 + \lambda\ \KL{q(\zi \mid \bi; \bsy{\theta}_e)}{p(\zi)} \right],
\label{eq:optim}
\end{equation}

with respect to $\bsy{\theta}$, where $\lambda = 2c$ can be thought of as a tuning parameter. Typically, $L$ is set to be small to encourage exploration of the objective function when optimising. As in \cite{kingma_2014}, we set $L = 1.$

\vskip10pt 

Once trained (i.e., estimates of $\bsy{\theta}$ are available), known sensitivity plumes, $\bs{b}_1, \bs{b}_2, \dots, \bs{b}_N,$ can be put through the encoder to obtain their corresponding variational means and variances on the latent space. The spatio-temporal emulator (Section \ref{sec:emulation}) is then used on this latent space to obtain predictions and prediction variances for new variational means, corresponding to $\bs{w}^*_j, j = 1, 2, \dots, M.$

\subsection{Emulation} \label{sec:emulation}

Here, we describe the way we use the information gathered from feature extraction of $\bs{B}$ (Section \ref{sec:feat}) to build an emulator that can predict plumes at new spatio-temporal locations, which can then be used to construct the matrix $\widehat{\bs{B}}^*$. With a slight abuse of notation, denote the $r$ coefficients for the features extracted for the $i$th plume by one of the two methods in Section \ref{sec:feat} as $\bs{u}_i \equiv (u_{i, 1}, u_{i, 2}, \dots, u_{i, r})',$ where $i = 1, 2, \dots, N$ (these would correspond to either rows of the matrix $\bs{U}$ for the EOFs, or variational means for the CVAE). We describe below the emulators we use to predict the elements of each coefficient in $\bs{u}^*_j$, at a new spatio-temporal location $\bs{w}^*_j$, for $j = 1, 2, \dots, M.$

\vskip 10pt

Let $\bs{W} \equiv (\bs{w}_1, \bs{w}_2, \dots, \bs{w}_N)'$, and collect the coefficients of feature $\gamma$ of the plumes in $[\bs{u}]_{\gamma}$; that is, define $[\bs{u}]_\gamma \equiv (u_{i,\gamma}: i = 1,\dots,N)'$. We use independent Gaussian process emulators \cite[][Chapter 2]{rasmussen_2006}, so that the prediction of the coefficient of feature $\gamma$ for a plume at $\bs{w}^*_j$ is normally distributed with mean 

\begin{equation}
	S_\gamma(\bs{w}^*_j, \bs{W}) S_\gamma(\bs{W}, \bs{W})^{-1} [\bs{u}]_\gamma, \quad \gamma = 1, 2, \dots, r, %\bs{W}_{n, J}(\tilde{\bs{w}}_i),
	\label{eq:GPmean}
\end{equation}

and variance 

\begin{equation}
	S_\gamma(\bs{w}^*_j, \bs{w}^*_j) - S_\gamma(\bs{w}^*_j, \bs{W})S_\gamma(\bs{W}, \bs{W})^{-1}S_\gamma(\bs{W}, \bs{w}^*_j), \quad \gamma = 1, 2, \dots, r,
	\label{eq:GPcov}
\end{equation}

where $S_\gamma(\cdot,\cdot)$ is the spatio-temporal squared-exponential kernel. Specifically, for two spatio-temporal points $\bs{w}_a = (\tilde{\bs{s}}_a'; t_a)'$ and $\bs{w}_b = (\tilde{\bs{s}}_b'; t_b)',$

\[S_\gamma(\bs{w}_a, \bs{w}_b) = \sigma_{st, \gamma}^2 \exp\left(- \frac{\Vert\tilde{\bs{s}}_a - \tilde{\bs{s}}_b\Vert_2^2}{2l_{s, \gamma}} - \frac{\vert t_a - t_b\vert}{2l_{t, \gamma}}\right)^2, \qquad \sigma^2_{st, \gamma}, l_{s, \gamma}, l_{t, \gamma} \in \mathbb{R}^+,\]

where $l_{s, \gamma}$, $l_{t, \gamma}$, and $\sigma^2_{st, \gamma}$ denote the spatial length scale, the temporal length scale, and the variance, of the Gaussian process for feature $\gamma$, respectively. We obtain estimates for $\bs{u}^*_j$, denoted $\widehat{\bs{u}}^*_j$, via \eqref{eq:GPmean}, and the corresponding covariance matrix via \eqref{eq:GPcov}. This gives us a Gaussian distribution over each emulated feature. We then draw samples from this distribution and pass them through the decoder in order to obtain a Monte Carlo approximation of the predictive distribution of the sensitivity plume at $\bs{w}_j^*$. 

\section{Results \& Discussion}

In this section, we present and discuss the results obtained when applying our methodology described in Section \ref{sec:methodology} to sensitivities computed from LPDMs.

\subsection{Dimension reduction and reconstruction}
\label{sec:results-1}

To train the CVAE and extract the EOFs, we obtained 20000 sensitivity plumes from individual simulation runs using FLEXPART. In each simulation, 2000 particles were released from a spatio-temporal location, and their trajectories followed backward in time for 30 days. Five thousand release locations were randomly chosen from within each of four spatio-temporal regions. Releases occurred over West Europe, Australia, North-West America, and the United Kingdom/Ireland, though the simulation region in each case was larger to allow plume tails to be captured. The spatial domain for each simulation was a $128 \times 128$ grid of resolution $0.352^\circ \times 0.234^\circ,$ with boundaries described in Table \ref{tab:FP-regions}. At regular temporal intervals over the 30 days, the length of time which particles spent within the first 100 m above ground level was recorded for each grid cell, in units of m$^3$ s g$^{-1}$ \citep{seibert_2004}. We then converted these quantities to time-integrated sensitivities (sometimes referred to as footprints, e.g., \cite{oney_2015}) in units of ns g$^{-1}$ (see Appendix \ref{app:sensitivity_calcs} for details). Finally, any sensitivity plumes deemed to have weak signal (defined as plumes with at most 10 grid cells with a sensitivity above the lowest 99.5\% of all sensitivities across all simulations) were removed from the dataset. To minimise spatial variability between all plumes as much as possible before training, we then subracted the location of each plume's origin from all $\bs{s}_k, k = 1, 2, \dots, K$, and rotated each plume such that its departure angle was 0$^\circ$ (the horizontal (East-West) axis in our images). This required us to estimate the original departure angle of each plume (details on how this was done are given in Appendix \ref{app:dep-angle-est}). After rotation, inverse distance weighting was used to reduce the resolution of the images down to $64 \times 64$, to help reduce the computational burden when training the CVAE. 

\begin{table}[t!]
	\caption{\doublespacing Details of the four regions over which FLEXPART simulations were run. Spatio-temporal locations for simulations were chosen on a subset of the simulation region in order to capture plume tails. Latitude is given in degrees North, longitude in degrees East, and time is of the format YYYY-MM-DD hh:mm:ss UTC.}
	\label{tab:FP-regions}
	\begin{center}
		\begin{tabular}{|c|cc|cc|cc|}
			\hline
			\textbf{Region} & \textbf{Min lat} & \textbf{Max lat} & \textbf{Min lon} & \textbf{Max lon} & \textbf{Min time} & \textbf{Max time} \\
			%& latitude & latitude & longitude & longitude & (YYYY-MM-DD & (YYYY-MM-DD \\
			%& (degrees) & (degrees) & (degrees) & (degrees) &  &  \\
			\hline 
			%\hline 
			West Europe & 46.000 & 75.952 & $-$3.000 & 42.056 & 2011-02-02 & 2011-07-01 \\
			& & & & & 00:00:00 & 00:00:00 \\
			\hline 
			Australia & $-$40.000 & $-$10.048 & 112.000 & 157.056 & 2012-09-02 & 2012-12-31 \\
			& & & & & 00:00:00 & 00:00:00 \\
			\hline 
			North-West & 30.000 & 59.952 & $-$116.000 & $-$70.944 & 2013-02-02 & 2013-07-01 \\
			America & & & & & 00:00:00 & 00:00:00 \\
			\hline 
			United Kingdom & 36.000 & 65.792 & $-$15.000 & 30.056 & 2014-09-02 & 2014-12-31 \\
			\& Ireland & & & & & 00:00:00 & 00:00:00 \\
			\hline
		\end{tabular}
	\end{center}
\end{table}

\vskip 10pt

\begin{figure}[t!]
	\centering \includegraphics[scale = 0.45]{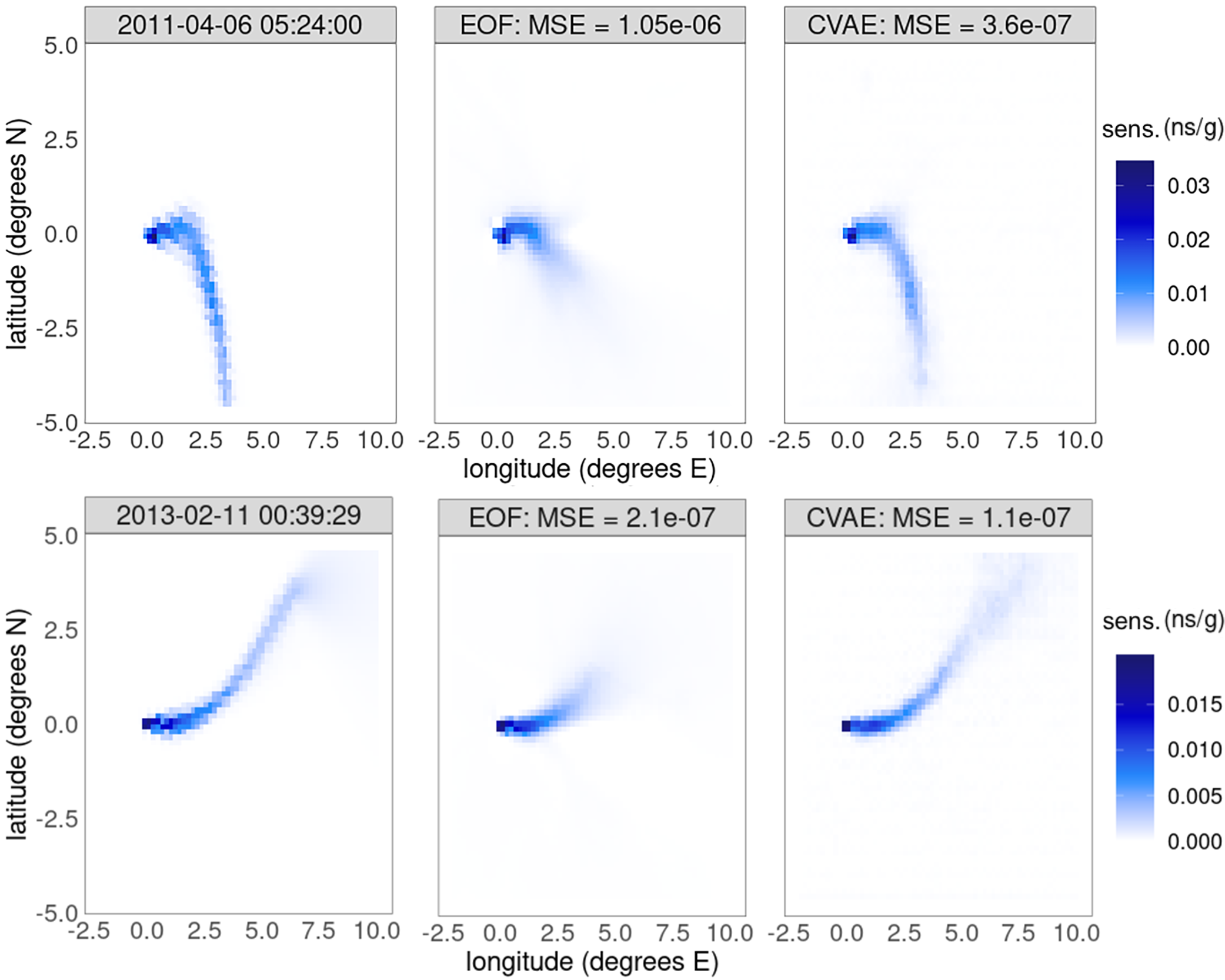}
	\caption{Left: Rotated and shifted sensitivity plumes as produced by FLEXPART, at the indicated time and date. Middle: EOF reconstruction of the same plumes, using 20 singular values, eigenvectors, and their corresponding coefficients. Right: CVAE reconstruction of the same plumes, using a 20-dimensional latent space. The MSE (in units of ns$^2$ g$^{-2}$) is shown above each reconstruction.}
	\label{fig:recon1}
\end{figure}

The EOFs were obtained using 70\% (11470) of the filtered set of simulated FLEXPART plumes. We retained the first $r = 20$ singular values ($\bs{D}_{20}$) and eigenvectors ($\bs{V}_{20}$), along with their corresponding coefficients ($\bs{U}_{20}$). To assess the error due to dimensionality reduction, we computed a mean squared error (MSE) for each plume by averaging the sum of the differences between the true values and the reconstructed values of the sensitivities at each grid cell. As shown in the top-left cell of Table \ref{tab:mses}, the sum of MSEs (sumMSE) for all plumes after EOF reconstruction was 0.00697 ns$^2$ g$^{-2}$. The middle panel in both rows of Figure \ref{fig:recon1} shows the reconstruction from the first 20 EOFs of two of the plumes used to obtain the EOFs. As can be seen, particularly in the second case, the tail end of the plume is not well captured by the EOF reconstruction.

\begin{table}
	\caption{\doublespacing The sumMSEs (in units of ns$^2$ g$^{-2}$) between the plumes and their reconstructions. The FLEXPART sumMSEs are calculated after reducing the dimension of the plumes down to $r = 20$, and then reconstructing them again, either using EOFs or via the CVAE. The NAME sumMSEs are calculated after emulation, rotation, and translation.}
	\begin{center}
		\begin{tabular}{lcc}
			\hline
			& \textbf{FLEXPART (Training)} & %\textbf{NAME (rotated space)} & 
			\textbf{NAME (Application)} \\
			\hline
			EOF & 0.00697 %& 0.000953 
			& 0.000312 \\
			CVAE & 0.00274 %& 0.000912 
			& 0.000260 \\
			\hline
		\end{tabular}
	\end{center}
	\label{tab:mses}
\end{table}

\vskip 10pt

\begin{figure}[t!]
	\centering \includegraphics[scale = 0.18]{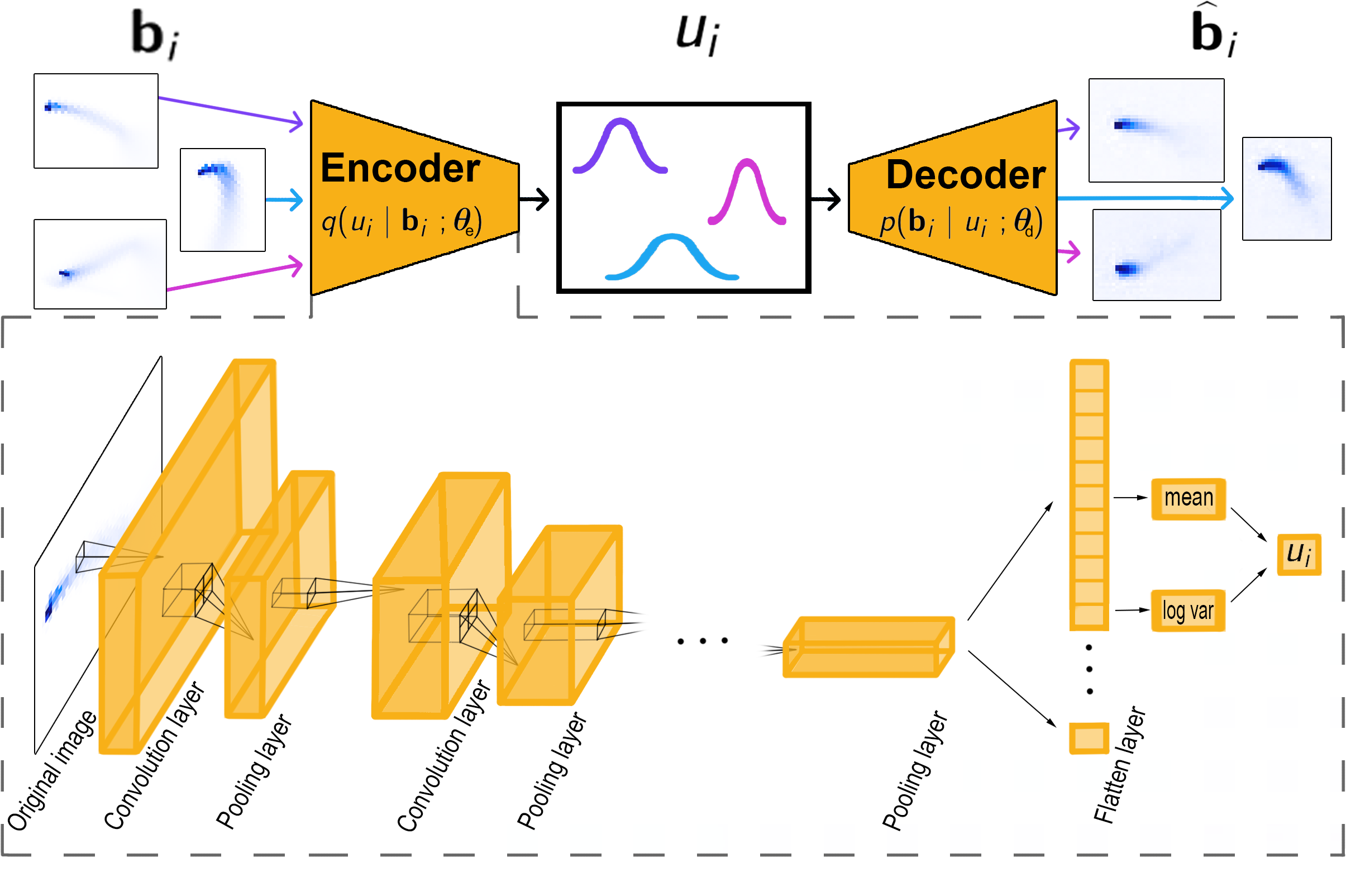}
	\caption{Visual representation of the CVAE architecture used in Section \ref{sec:results-1}. The vector $\bs{b}_i$ represents the true (in practice, rotated and shifted) plumes, while $\widehat{\bs{b}}_i$ represents the reconstructed plumes. In this figure, $u_i$ is a one-dimensional latent variable. We also show the make-up of the encoder: convolutional layers followed by max-pooling layers and a flattening layer. The decoder is made up similarly with convolutional 2D transpose layers.}
	\label{fig:cvae}
\end{figure}

The CVAE was trained using the same 70\% of simulated FLEXPART plumes as used to obtain the EOFs. The remaining 30\% were used for validation during training, and we set the latent dimension to $r = 20$. The encoder portion of the CVAE architecture consisted of six 2D convolutional layers with max pooling and scaled exponential linear unit (selu) activation functions, a flattening layer, and then two dense layers with leaky rectified linear unit (leaky relu) activation functions, one for the latent mean vector, and one for the latent log-variance vector. The decoder portion of the CVAE architecture consisted of a single dense layer, followed by a layer which reshaped the variables into 2D images, and then six 2D convolutional transpose layers with selu activation functions. A schematic of the full architecture is shown in Figure \ref{fig:cvae} for a latent space of dimension $r =1$. Using cross-validation, we found that our model was largely insensitive to small values of $\lambda$, and that it underfit for large $\lambda$. This suggests that it is relatively easy to estimate the mean and variance encoding maps, as well as the mean decoder map, from the data, and that not much regularisation is needed in our case. We therefore fixed $\lambda$ to $10^{-9}$; note that a small value for $\lambda$ does not mean the resulting network is a standard autoencoder since each $\zi$ is still associated with some non-trivial variational mean and variance in $q\left(\zi \mid \bi; \bsy{\theta}_e \right)$. Training was done using batches constructed from random samples without replacement of inputs of a fixed size ($< N)$ from the full training data set. The CVAE objective shown in $\eqref{eq:optim}$ was then computed by summing over the batch rather than the full data set. This procedure was repeated until the full training set was sampled several times (a single run through the entire data set is known as an \textit{epoch}). Parameters and weights were randomly initialised, and variables were randomly sampled on the latent space from distributions defined via the encoded means and log-variances throughout training (see \eqref{eq:MC}). Due to the complexity of the objective function and the sensitivity of the parameter estimates to initial conditions, we elected to train the network 10 times over 500 epochs using an adaptive moment (Adam) optimiser. The trained network with the lowest training MSE was then chosen for emulation.

\vskip 10pt 

Occasionally, our dimension-reduction procedures yielded negative sensitivities for some grid cells after reconstruction. For example, one randomly selected plume from the CVAE yielded 90 negative sensitivities out of a total of 4096, with the minimum and maximum sensitivities being $-$0.0013 ns g$^{-1}$, and 0.052 ns g$^{-1}$, respectively. Another randomly chosen plume yielded 6 negative sensitivities, with the minimum and maximum sensitivities being $-0.00084$ ns g$^{-1}$, and 0.018 ns g$^{-1}$, respectively. Negative sensitivities do not make physical sense, and we therefore set all negative sensitivities (from both dimension-reduction procedures), when these arose after reconstruction, to zero. The right panels in both rows of Figure \ref{fig:recon1} show the resulting CVAE reconstructions for two training plumes after truncation. When compared to the EOF reconstructions, we see that the CVAE is able to retain more of the important structural information about the plumes during the dimension reduction stage. This is further supported by the sumMSE value shown in the bottom-left cell of Table \ref{tab:mses} (0.00274 ns$^2$ g$^{-2}$), which is less than half the sumMSE for the EOFs.

\vskip 10pt

Having demonstrated the ability for the CVAE to reduce, and reconstruct, plumes with superior accuracy to the conventional EOF-based approach, we now turn our attention to emulation. We note that the methods we present for dimension reduction are not LPDM-dependent. That is, we do not need to use FLEXPART plumes, despite it having been the LPDM used to generate the data that the CVAE was trained on and that the EOFs were  obtained from. As we show next, our models are general enough that we can apply them to plumes from other LPDMs.

\subsection{Application}
\label{sec:results-2}

To demonstrate the utility of our surrogate plume models and subsequent emulation, we apply them to plumes simulated using the Numerical Atmospheric-Dispersion Modelling Environment (NAME), which were used for flux inversion in previous works (\citealp{ganesan_2015}, \citealp{zammit_2015}, \citealp{zammit_2016}). These plumes were simulated over the United Kingdom and Ireland every two hours between 1 January 2014 00:00:00, and 30 April 2014 22:00:00, using a $128 \times 128$ spatial grid of resolution $0.352^\circ \times 0.234^\circ.$ The particle trajectories were followed backward in time for 30 days. Each sensitivity plume originated at one of the four measurement sites shown in the top-left panel of Figure \ref{fig:plumes-corr}. 

\vskip 10pt

Every second plume was removed from the dataset, and those plumes that were removed were then emulated using solely those plumes that were retained. With the CVAE, those which were retained were passed through the encoder in order to find the corresponding variational means. The mean vectors corresponding to the plumes were then used to emulate mean vectors for those plumes we removed from the dataset. We did this by fitting GP emulators to each plume ``feature" (see Section~\ref{sec:emulation}); the parameters $l_{s, \gamma}$, $l_{t, \gamma}$, and $\sigma_{st, \gamma}$ in each case were fit via maximum likelihood estimation (MLE). We also used two additional GP emulators: one for the East-West component of the plume departure angle, and one for the North-South component. We emulated values for both these components for each of the plumes removed from the NAME dataset. In order to account for the uncertainty in the plume reconstruction and the rotation and translation back to the original space, we sampled 100 realisations from each Gaussian distribution on the latent space, and used these realisations to construct, and rotate, 100 emulated plumes to their predicted departure angles. The origins of the plumes were then shifted from $(0, 0)$ to the appropriate spatial location (i.e., the coordinates of one of the measurement sites), and inverse distance weighting was used to bring the spatial resolution back up to $128 \times 128$. Finally, we took the mean and standard deviation over the grid cells to obtain a mean plume and associated prediction standard errors, respectively. After these transformations, the sumMSE, as shown in the bottom-right cell of Table \ref{tab:mses}, was 0.000260 ns$^2$ g$^{-2}$. 

\vskip 10pt

For comparative purposes, the same analysis was performed using the EOFs. Using the existing matrices $\bs{D}_r$ and $\bs{V}_r$ found from the FLEXPART training dataset, we used linear regression to find the corresponding coefficients for the 50\% of plumes left in the NAME dataset. GP emulators were then fit to each coefficient and both angle components, with parameters again found via MLE. As before, to build an emulated plume, 100 samples from the distribution of each emulated coefficient were drawn. These samples were then combined with the existing singular values and eigenvectors to construct emulated plumes. The mean and variance over the grid cells were computed to construct a mean plume and the associated prediction standard errors. As shown in the top-right cell of Table \ref{tab:mses}, the sumMSE obtained was 0.000312 ns$^2$ g$^{-2}$, which is comparable, but slightly worse than what we obtained using the CVAE.

\vskip 10pt

\begin{figure}[t!]
	\centering \includegraphics[scale = 0.45]{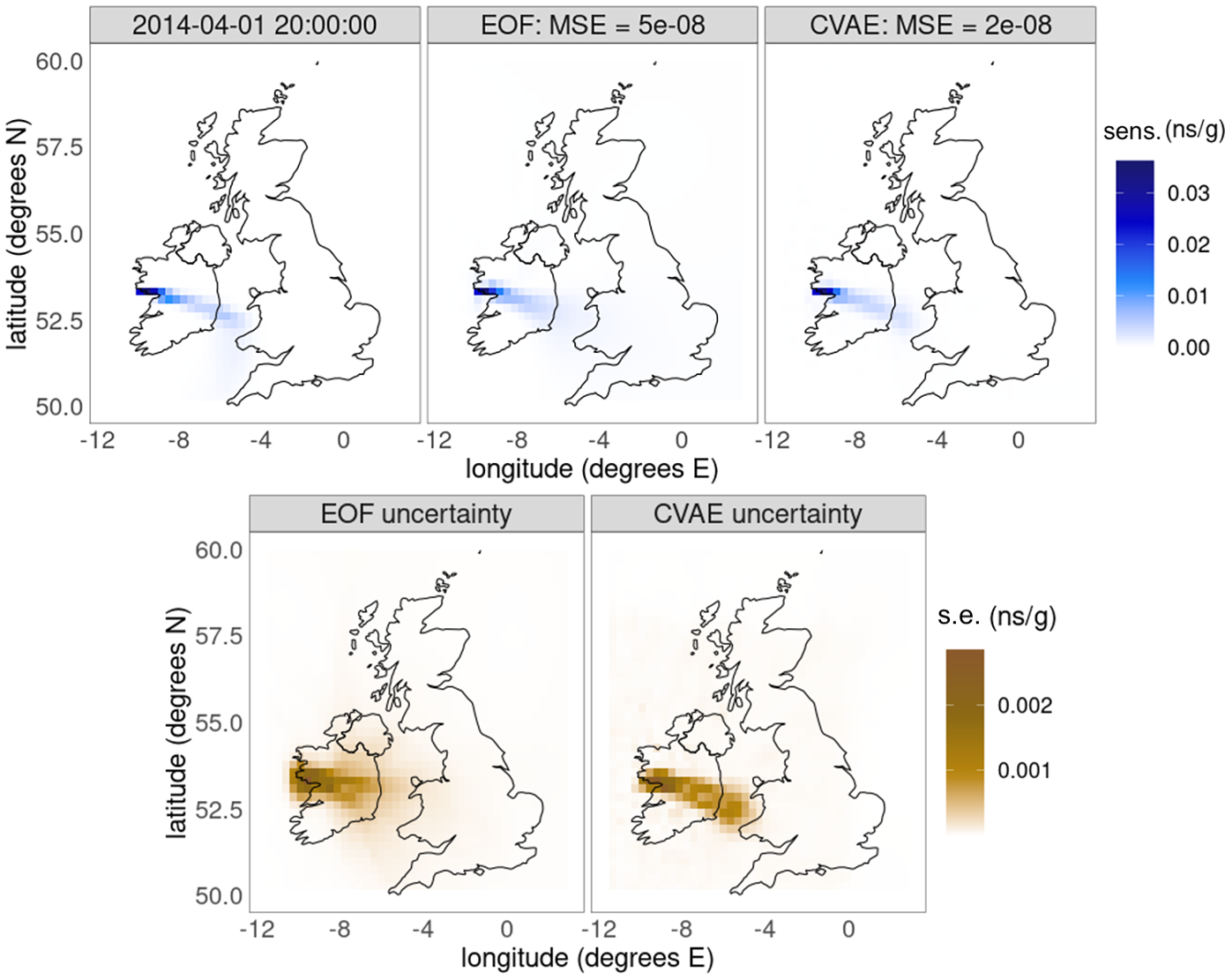} \\
	\caption{Top: Sensitivity plume as produced by NAME on 1 April 2014 at 20:00:00, followed by the emulated reconstruction using the EOFs, and the CVAE, respectively, along with their MSEs (in units of ns$^2$ g$^{-2}$). Bottom: Prediction standard errors for both the EOF and CVAE methods. The emulated plumes and prediction standard errors were obtained from 100 Monte Carlo samples from the conditional distributions following GP emulation. In each case, samples were used to construct, rotate, and translate an emulated plume.}
	\label{fig:uncertainty}
\end{figure}

Figure \ref{fig:uncertainty} shows the emulated plumes for one prediction location and time point, using both the CVAE and EOFs for dimension reduction. More examples are given in Appendix C. 

\section{Conclusion}

In this article we present a novel method for emulating LPDM sensitivity plumes using a CVAE. The method is not LPDM dependent, making it flexible across many applications in atmospheric science. We show that the CVAE is able to capture important structural information within a collection of FLEXPART plumes in a 20-dimensional latent space, and hence produce reasonable reconstructions of these plumes. We also show through GP emulation on a set of known NAME plumes that we are able to reasonably reconstruct additional new plumes given information from nearby, known plumes. The effectiveness of the CVAE in dimension reduction when compared to the use of EOFs is highlighted. We find that the CVAE outperforms the EOFs both visually and via sumMSE metrics. 

\vskip 10pt 

Our framework will allow researchers to save time when performing flux inversions, and allow for the generation of numerous, higher-resolution sensitivity plumes than what was previously possible due to computational constraints. It is difficult to precisely quantify the amount of computational time that could be saved; however, the computing resources needed for this study are revealing. Using all cores on a 64-core server, it took us approximately two days to obtain 3000 plumes using FLEXPART. This includes the time we needed to obtain meteorological data using the software \texttt{flex\_extract}, and the time needed to post-process the simulation output. This is already substantial, despite the fact that our simulations only involved computing hourly calculations based on the position of the released particles, and ERA-Interim meteorology data which is only at a six-hourly temporal resolution. Using higher-resolution meteorological data would slow down the simulations considerably. Further, if a smaller simulation time step is desired, then the number of released particles must be increased in a commensurate manner (P. Seibert, communication via FLEXPART mailing list, June 26 2020): Separate timing tests revealed that if sensitivities were needed at five-minute intervals instead of hourly intervals, the time taken for every individual simulation would increase eight-fold. On the other hand, setting up and predicting with our emulator framework described in Section \ref{sec:results-2}, took approximately 10 hours. About one hour was needed to perform MLE to optimise each $l_{s, j}, l_{t, j}, $ and $\sigma^2_j,$ and approximately nine hours were needed to perform the MC sampling, and plume rotations and translations with code that was not optimised for speed. The time taken to do the actual emulation was less than a minute. The key attraction of our framework is that the timings associated with our emulator are independent of those associated with the FLEXPART simulations: If the number of particles or the frequency of sampling were to be increased in FLEXPART (resulting in an increase in computational time), the time taken to obtain plumes via our methodology would remain the same. This framework thus presents a huge opportunity for large savings in computational time for more complex simulations.

\vskip 10pt

In future work, it would be interesting to see whether this methodology could be used for emulating plumes associated with column-averaged greenhouse-gas measurements, which is needed when doing inversions from remote-sensing data. If so, this methodology could drastically reduce the computational time associated with flux inversions using data from satellite missions, which retrieve data at many spatial locations and time points, and hence would require many sensitivity plumes if used in a flux inversion setup. It should be noted, however, that our approach may not be useful if plumes become less correlated in space and time as a result of rapidly varying meteorological factors. The utility of our approach to remote-sensing data and to space-time regions with highly variable meteorology are the subject of future research.

\section*{Acknowledgements}

LC acknowledges the support of the Australian Government Research Training Program Scholarship. AZ-M's research was supported by the Australian Research Council (ARC) Discovery Early Career Research 
Award (DECRA) DE180100203, and by Discovery Project DP190100180. NMD's research was supported by the ARC Future Fellowship FT180100327. For his help with the FLEXPART and \texttt{flex\_extract} installations, the authors thank Yi Cao. For provision of the NAME simulations, the authors thank Alistair Manning. For their discussions related to the research, the authors thank Anita Ganesan, Rachel Tunnicliffe, Dan Pagendam, and Michael Bertolacci. 

\section*{Code and data availability}

The code required to do the emulation described in this article are available at \url{https://github.com/Lcartwright94/LPDM-Sensitivity-Emulation}. Instructions for accessing and downloading the accompanying datasets are also available via this link. 

\section*{Conflict of interest statement}

The authors declare they have no conflict of interest.

\bibliographystyle{apalike}
\bibliography{references-APR,references-Ems,references-FP,references-VAE}

\newpage

\appendix

\section{Post-processing of FLEXPART output} \label{app:sensitivity_calcs}

In this appendix we describe the procedure used to convert the output obtained from backward-simulations using FLEXPART to time-integrated sensitivity plumes. 

\vskip 10pt

We first sum the output (in units of s m$^3$ kg$^{-1}$) over all time points, which leads to an approximate integrated sensitivity, sometimes known as the ``footprint", of the trace gas \citep{oney_2015}. We next divide by the volume of the grid cell in each instance to remove the dependence on the size of the grid cell, thus yielding the sensitivities in units of s kg$^{-1}$. From here, we scale the sensitivities by $10^{-3}$ to make them have units of s g$^{-1}$. 

\vskip 10pt 

When doing flux inversion, one would need to multiply these sensitivities by an emission rate. Doing this, we would obtain a mass mixing ratio (in units of g g$^{-1}$). To instead ensure we obtain a mole fraction after this multiplication, which is typically used for reporting measurements of atmospheric methane, we scale the sensitivities by the molar mass of “air” (28.9644 g mol$^{-1}$) and divide by the molar mass of methane (16.0425 g mol$^{-1}$). This results in sensitivities in units of s g$^{-1}$ mol (of methane) mol$^{-1}$ (of air), and a mole fraction in units of mol (methane) mol$^{-1}$ (air) after multiplication by the emission rate. Finally, we scale our sensitivities by $10^9$, yielding units ns g$^{-1}$ mol (methane) mol$^{-1}$ (air), and so that the resulting value after multiplication by an emission rate is in units of nmol (methane) mol$^{-1}$ (air), or ppbv. 

\vskip 10pt 

For simplicity throughout this paper, we refer to the sensitivities as having units ns g$^{-1}$, rather than ns g$^{-1}$ mol mol$^{-1}$.

\section{Estimation of plume departure angle} 
\label{app:dep-angle-est}

\begin{figure}[ht!]
	\centering \includegraphics[scale = 0.3]{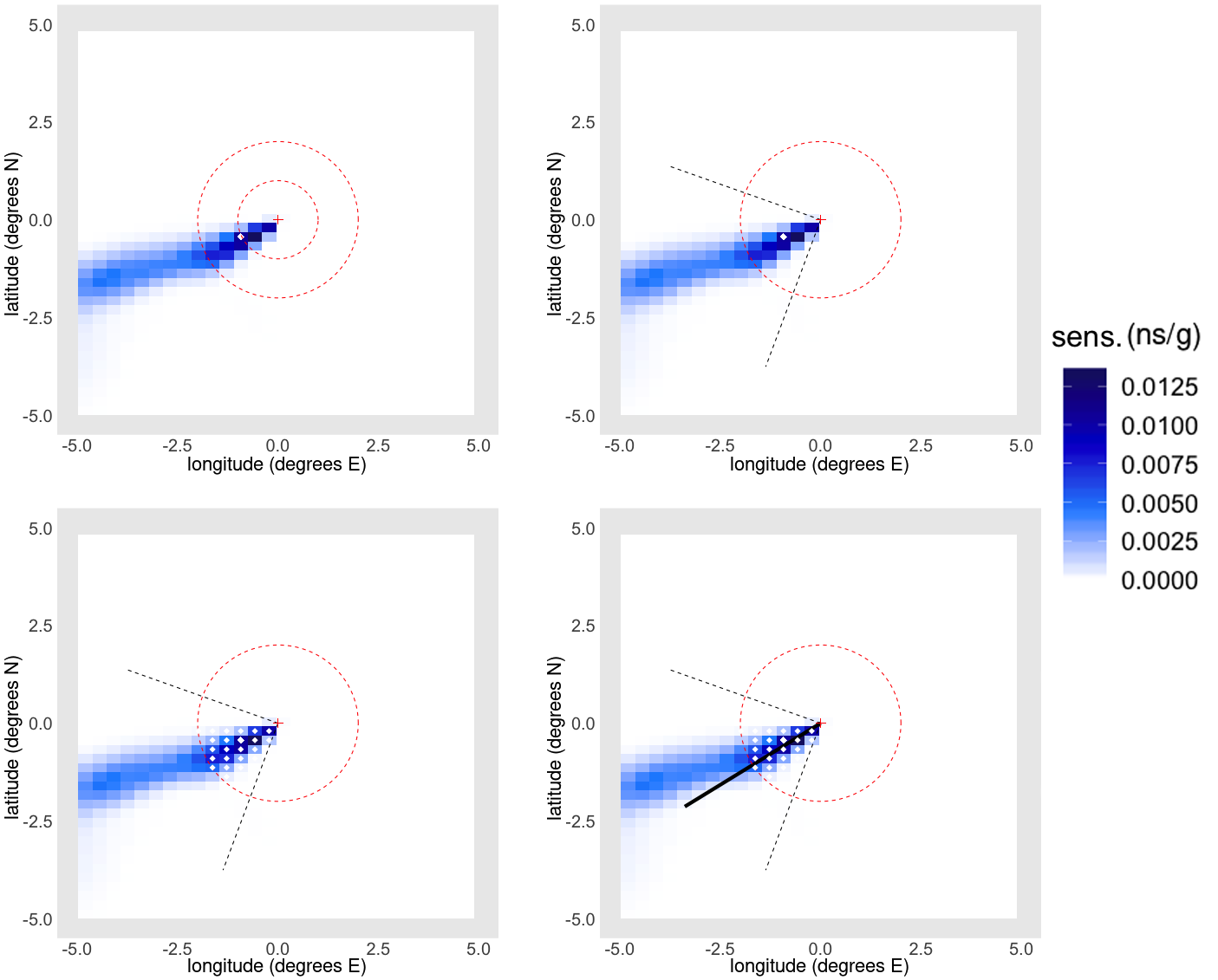}
	\caption{A visual representation of the departure-angle estimation process. Top-left: Step 1 -- Locating the maximum sensitivity within the annulus. Top-right: Step 2 -- Construction of the ``angle window". Bottom-left: Step 3 -- Locating all points within the window, and within two lat-lon degrees of the plume's origin. Bottom-right: Step 4 -- Angle determined by computing a weighted average of the points identified in the bottom-left.}
	\label{fig:angle-estimate}
\end{figure}

This appendix describes the steps taken to estimate the departure angle for each plume. Figure \ref{fig:angle-estimate} provides a visual representation of some of these steps.

\vskip 10pt 

First, we determine the approximate direction for the departure angle. This is done by constructing a small annulus around the plume's point of origin, and then computing the angle, $a$, subtended from the plume's origin to the grid cell with maximum sensitivity within the annulus. The reason an annulus was used, rather than a disc, was due to the occasional non-zero sensitivities located directly behind the plume's point of origin, particularly when turbulent weather was present. In such cases, a search within a fixed radius may lead to an estimated departure angle that is in the opposite direction to the true departure angle.

\vskip 10pt 

Following this, we construct an ``angle window"  $(a - \pi / 4, a + \pi / 4)$, and then compute a weighted average of the angles subtended from the plume's origin, where the weights are all sensitivities that are within this window and up to two lat-lon degrees away from the plume's origin. This weighted average is then our estimated plume departure angle. 

\section{Additional emulation results}\label{app:more-emulation-results}

This appendix shows an additional three sets of emulated plumes together with associated prediction standard errors. 

\begin{figure}[t!]
	\centering \includegraphics[scale = 0.4]{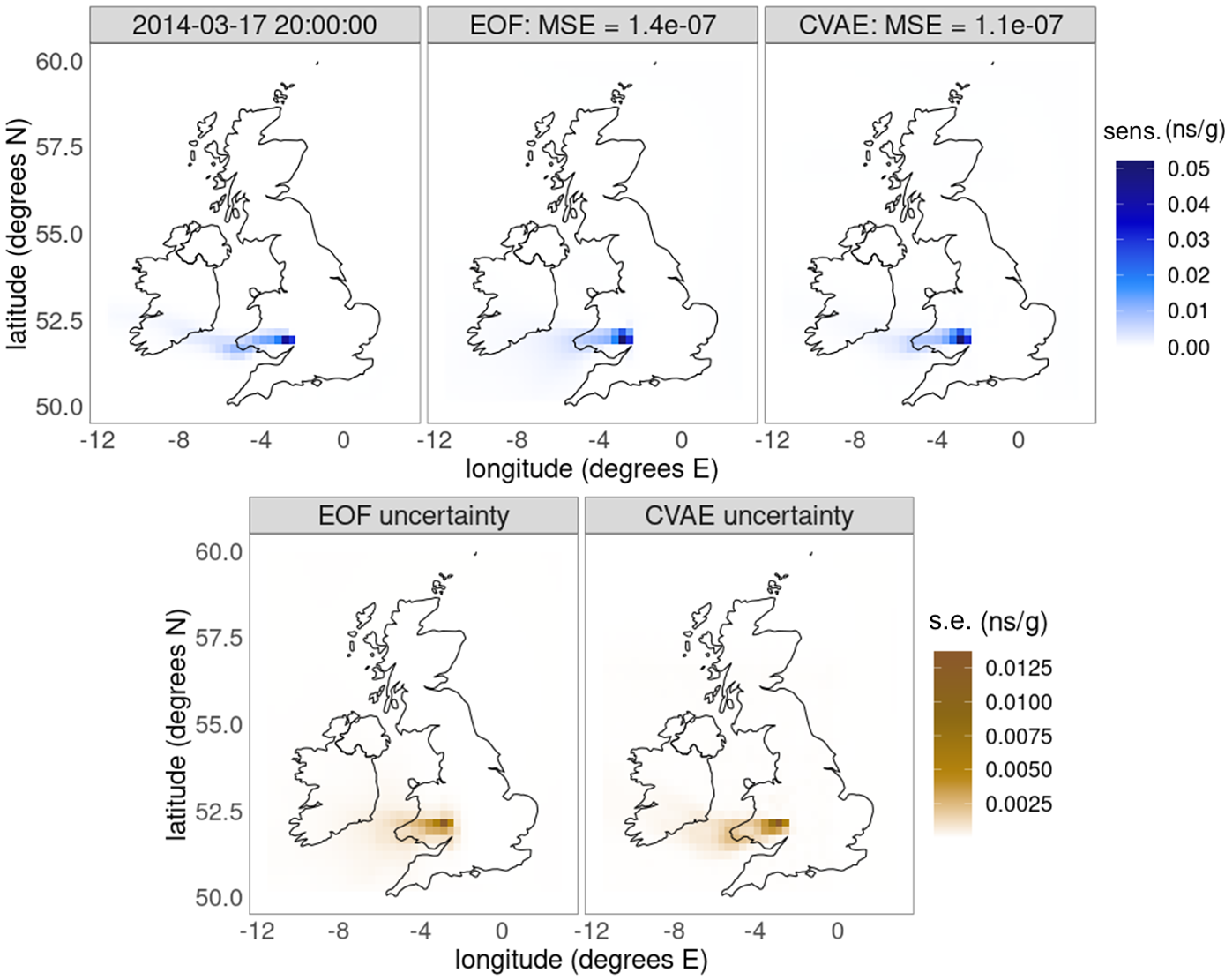}
	\caption{Same as Figure \ref{fig:uncertainty}, but for 17 March 2014 at 20:00:00.}
	\label{fig:app}
\end{figure}

\begin{figure}[t!]
	\centering \includegraphics[scale = 0.4]{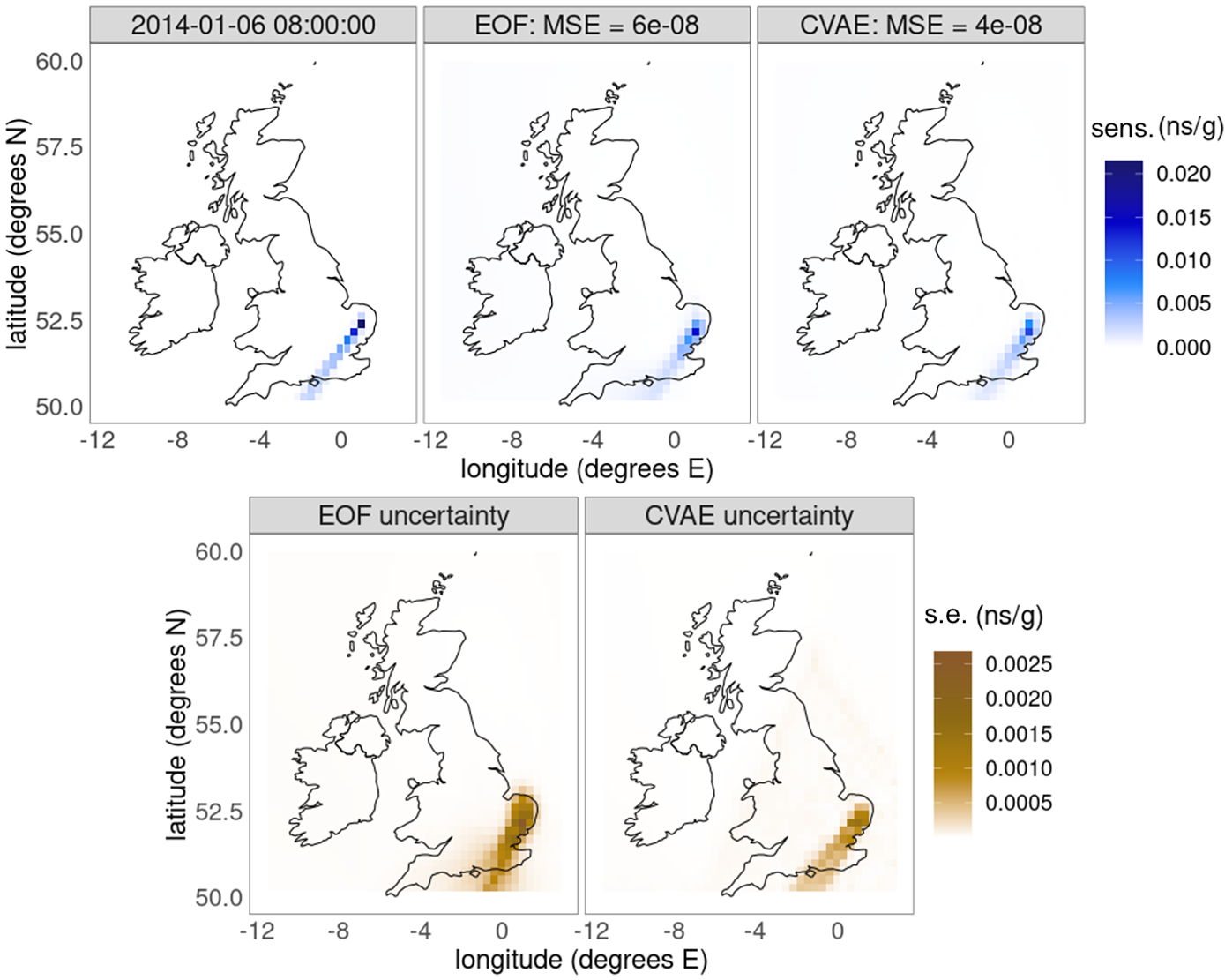}
	\caption{Same as Figure \ref{fig:uncertainty}, but for 6 January 2014 at 08:00:00.}
\end{figure}

\begin{figure}[t!]
	\centering \includegraphics[scale = 0.45]{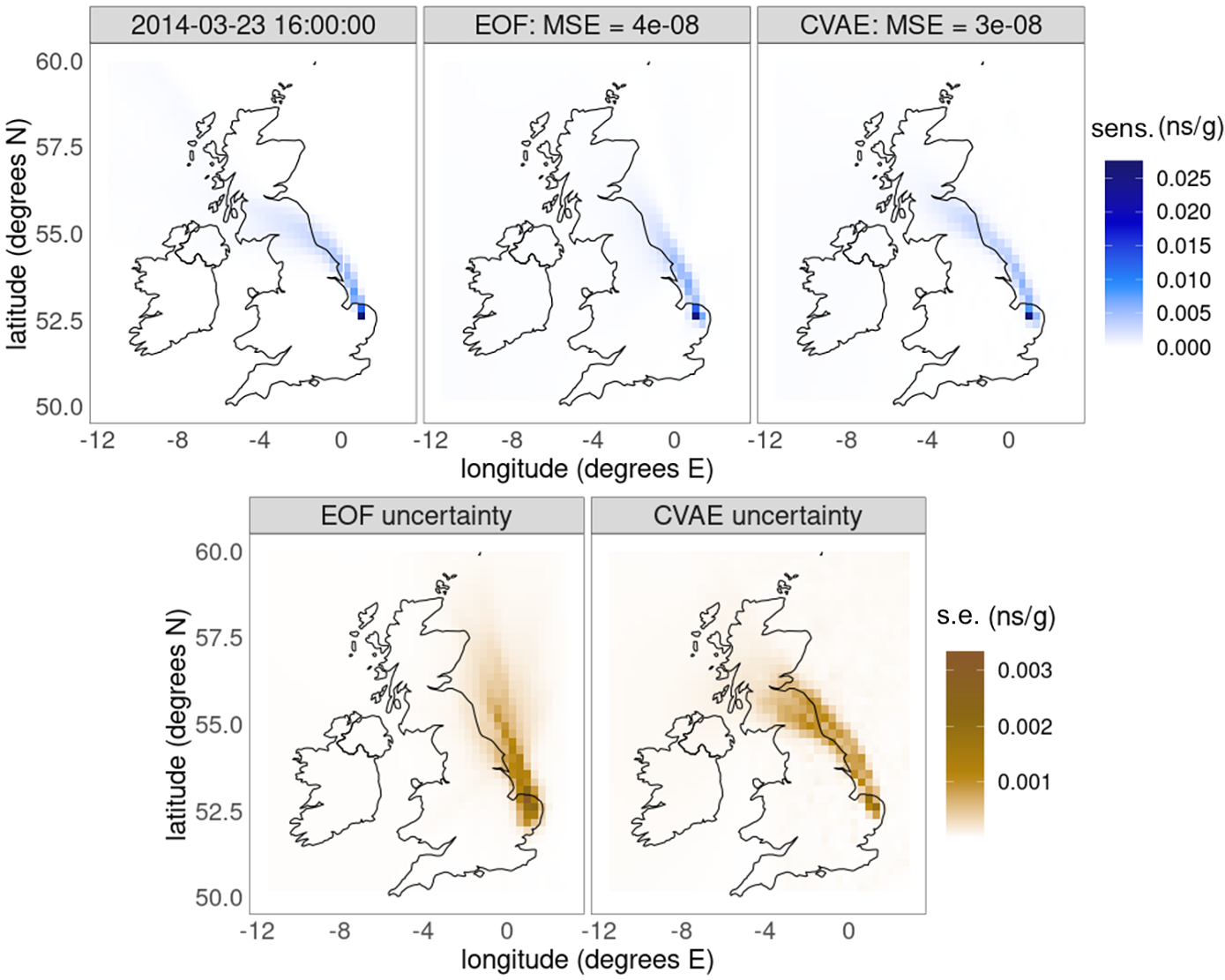}
	\caption{Same as Figure \ref{fig:uncertainty}, but for 23 March 2014 at 16:00:00.}
\end{figure}

\end{document}